\documentclass[10pt]{iopart}

\usepackage{iopams}  

\usepackage{microtype}
\usepackage{graphicx}
\usepackage{subfigure}
\usepackage{booktabs} 
\usepackage{natbib}
\bibpunct[, ]{(}{)}{,}{a}{}{,}%

\usepackage{hyperref}

\usepackage{algorithm}
\usepackage{algpseudocode}

\begin{document}

\title[Towards a Phenomenological Understanding of Neural Networks: Data]{Towards a Phenomenological Understanding of Neural Networks: Data}

\author{Samuel Tovey}
\address{Institute for Computational Physics, University of Stuttgart, Stuttgart, Germany}
\ead{stovey@icp.uni-stuttgart.de}

\author{Sven Krippendorf}
\address{Universit\"{a}ts-Sternwarte and Arnold Sommerfeld Center for Theoretical Physics, Faculty of Physics, LMU, Munich, Germany}
\ead{sven.krippendorf@physik.uni-muenchen.de}

\author{Konstantin Nikolaou}
\address{Institute for Computational Physics, University of Stuttgart, Stuttgart, Germany}

\author{Christian Holm}
\address{Institute for Computational Physics, University of Stuttgart, Stuttgart, Germany}

\begin{abstract}
A theory of neural networks (NNs) built upon collective variables would provide
scientists with the tools to better understand the learning process at every stage.
In this work, we introduce two such variables, the entropy and the trace of the empirical
neural tangent kernel (NTK) built on the training data passed to the model.
We empirically analyze the NN performance in the context of these variables
and find that there exists correlation between the starting entropy, the trace
of the NTK, and the generalization of the model computed after training is complete.
This framework is then applied to the problem of optimal data selection for the training of
NNs.
To this end, random network distillation (RND) is used as a means of selecting training data
 which is then compared with random selection of data.
It is shown that not only does RND select data-sets capable of
 outperforming random selection, but that the collective variables associated with the
RND data-sets are larger than those of the randomly selected sets.
The results of this investigation provide a stable ground from which the selection of
data for NN training can be driven by this phenomenological framework.
\end{abstract}


\ioptwocol

\section{Introduction}\label{sec:introduction}
Neural Networks (NNs) are a powerful tool for tackling an ever-growing list of
data-driven challenges.
Training NNs is a problem of model fitting over a parameter space so large (in some cases infinite~\cite{rasmussen05a})
that in their finite width regimes they are powerful feature learning devices and in
their infinite regimes, regression-driven universal approximators of functions~\cite{hornik89a}.
These methods have experienced terrific success in both day to day technology including
speech recognition, tailored advertising, and medicine as well as many scientific fields.
Whilst theoretical methods have been making steady headway into understanding
the processes underlying machine learning, what is still absent is a simple,
physically inspired, phenomenological framework  to understand NN training.
That is, a model that describes the learning process independent of microscopic variables
that go into the training and deployment, ideally motivated by well studied physical principles.
These variables include model complexity defined by the number of layers, layer width, and
propagation algorithm used by the neural network, the data-set used to train the model such as the size of
the set used or its coverage of the problem space, and finally the algorithms used to
minimize the chosen loss function and train the NN such as the optimizer or even the loss
function itself.
In this work, NN performance is analyzed in terms of the initial state of the
empirical neural tangent kernel (NTK) (see~\ref{subsec:neural-tangent-kernel}).
Use of the NTK arises here naturally as it holds crucial information on training
dynamics including both the NN and the data on which it is trained.
With this approach, we are interested in a universally calculable set of variables from the NTK which can be
used to analyse NN behaviour across data-sets and architectures.
As the spectrum of the NTK has been observed several times to be sparse (i.e dominated by
a single eigenvalue and smaller ones, the vast majority of which are 0),
it seems feasible to compress the information in the NTK down to a few collective
variables, in this case, the trace of the NTK and the entropy computed from its eigenvalues.
Such a framework should allow us to optimise the training process.
To do this, we identify how these variables are related with training performance
(e.g. generalisation error) and then use this information to optimise NN training.
In particular, here this framework is applied to the problem of data-selection for the
training of NNs.
Namely, random network distillation (RND) is examined as a method that constructs
data-sets for which the collective variables are larger than that of a randomly selected set,
resulting in improved generalization.
Novel observations include correlation between the starting entropy and trace
of the NTK of a NN with model performance as well as insight into why RND is so performant.
The results presented here provide a clear path for future investigations into the construction
of a phenomenological theory for machine learning training built upon foundations
in physically motivated collective variables.

\subsection{Related Work}\label{subsec:related-work}
Research into the NTK has exploded in the last decade.
As such, several groups have made promising steps in directions somewhat aligned with
the work presented here.
Kernel methods as applied to NNs were introduced as far back as the 1990s
when initial results were found on the relationship between infinite width NNs
and Gaussian processes~\cite{neal95a} (GPs).
Since then, focus has shifted towards an alternative kernel representation of NNs,
namely, the NTK.
The theory developed in this paper is built upon the empirical NTK, that is, the NTK
matrix computed for a finite size NN on a fixed data-set.
Work on the NTK first appeared in the 2018 paper by~\citet{jacot18a} where it was introduced
as a means of understanding the dynamics of an NN during training as well as
to better characterize their limits.
Since then, the NTK has been used as a launching pad for a large number of investigations
into the evolution of NNs.
In the direction of eigenspectrum analysis, \citet{ari18a} describes the splitting of
the eigenvectors of the Hessian during training and how this affects gradient descent.
Their research shows that the gradients converge to a small subspace spanned by a set of
eigenvectors of the Hessian, the dimension of which is determined by the problem complexity, e.g,
the number of classes in a data-set.
In their 2021 paper, \citet{ortiz21a} extend the work of \citet{ortiz20a} by further discussing
the concept of neural anisotropy directions (NADs) and how they can be used to explain what
 makes training data optimal.
They find that NNs, linearized or not, sort complexity in a similar way
using the NADs.
Further, they draw upon foundations in kernel theory, specifically that the complexity
of a learning problem is bound by the kernel norm chosen for the task.
This reduces to stating that the goal of a learning model is to fit the eigenfunctions
of the kernel.
When applying this to NNs, they discovered that NNs struggle to
learn on eigenfunctions with small associated eigenvalues.
Whilst these studies have aimed to characterize the NTK in its static form, some prior
work has been done on understanding the evolution of this kernel during training.
In their 2022 paper, \citet{krippendorf22a} demonstrated a duality between cosmological
expansion and the evolution of the NTK trace throughout training.
Mathematically, this involved re-writing NN evolution as a function of the
eigenvalues of the NTK, a formulation drawn upon in this work to highlight the role of our
collective variables.
Of the above examples, all are directly related to NNs.
However, most NN theory finds its foundations in kernel theories as they have been
thoroughly studied and allow for exact solutions.
It has been long-established in the kernel regression community that the use of maximum
entropy kernels can provide fitting models with the best base from which to fit.
The concept of these kernels was first described by \citet{tsuda04a} wherein they demonstrate
that the diffusion kernel is built by maximising the von-Neumann entropy of a data-set.

This work aims to extend the previous studies presented here to finite-size NNs
with a focus on the data-selection process.
The remainder of the paper is structured as follows.
In the next section, the theoretical background required to understand the collective
variables is developed.
Following this, two experiments are introduced and their results discussed.
The first of these experiment involves understanding the role of the collective
variables in model training.
The second looks to using these collective variables to explain why RND data selection
outperforms randomly selected data-sets.
Finally, an outlook of the framework is presented and future work discussed.

\section{Preliminaries}
\label{sec:preliminaries}
Throughout this work, several important concepts related to information theory, machine
learning, and physics are relied upon.
In this section, each of these concepts is introduced and explained such that the use of
our collective variables is motivated.

\subsection{Neural Tangent Kernel}
\label{subsec:neural-tangent-kernel}
The NTK came to prominence in 2018 when papers began to arise demonstrating
analytic results for randomly initialized, over-parametrized, dense NNs~\cite{jacot18a, lee20a}.
This research resulted in the demonstration that in the infinite width limit, NNs
evolved as linear operators and provided a mathematical insight into the training dynamics.
This has since been extended so that it is applicable to most NN architectures and work
is currently underway towards better understanding the learning mechanism in this regime~\cite{yang19a}.
Given a data-set $\mathcal{D}$, the NTK, denoted $\Theta$, is the Gramian matrix~\cite{horn90a} formed by
the inner product
\begin{equation}
    \centering
     \Theta_{ij} = \sum\limits_{k} \frac{\partial}{\partial \theta_{k}}f(x_{i}, \{\theta\})\cdot \frac{\partial}{\partial \theta_{k}}f(x_{j}, \{\theta\})~,
    \label{eq:ntk-componant}
\end{equation}
where $\Theta_{ij}$ is a single entry in the NTK matrix, $f$ is an NN with a single output dimension,
$x_{i} \in \mathcal{D}$ is a data point, and $\{\theta\}$ are the parameters of the
network $f$.
Individual entries in the NTK matrix provide information about how the representation
of a point $x_{i}$ will evolve with respect to another $x_{j}$ under a change of parameters,
that is, it is  an inner product between the gradient vectors formed by the
NN  representations of data in the training set with respect to the
current state of the network.
The role of the NTK matrix in NN training is best described in the infinite-width limit by the continuous
time update equation for the representation of a single datapoint by an NN, which,
under gradient descent, may be written
\begin{equation}
    \centering
    \dot{f}(x_{i}) = -\sum\limits_{x_{j}}\Theta_{i, j}\frac{\partial \mathcal{L}(f(x_{i}, \{\theta\}), y_{i})}{\partial f_{x_{j}, \{\theta\}}}~,
    \label{eq:ct-ntk-update}
\end{equation}
where $\mathcal{L}$ is the loss function and $y_{i}$ is the
label for the $i^{{\rm th}}$ element of the training data-set corresponding to input
point $x_{i}$~\cite{lee20a}.

\subsection{NTK Spectrum}
\label{subsec:ntk-spectrum}
In the infinite-width regime, the NTK built on a data-set is constant throughout training
and therefore, may be used as an operator to step through model updates~\cite{jacot18a}.
However, in finite network regimes, such as those widely deployed in science and industry,
this is not the case and the NTK will evolve as the model trains.
In this work, the state of the NTK before training becomes a measurement device
for understanding how a model will generalize.
Therefore, it is crucial to have the correct tools with which to discuss and quantify this
matrix.
The tools used in this investigation trace their roots to random matrix theory,
information theory, and physics, beginning with entropy.
The Shannon entropy, $S^{Sh}$, describes the amount of information contained within a
random variable~\cite{shannon48a} and can be written
\begin{equation}
    S^{Sh} = -\sum\limits_{i} p(x_{i})\ln p_{i}(x_{i}),
    \label{eq:shannon-entropy}
\end{equation}
where $p(x_{i})$ is the probability of random variable $x_{i}$ being realised, and $\ln$ denotes a logarithm.
In his original work, and as is still common in information theory today, a log of
base 2 was chosen due to the limited domain of binary numbers.
For the purpose of this work, we use the natural logarithm as the variables take on
continuous values.
Von Neumann entropy arose in the field of quantum mechanics upon the introduction of
the density matrix as a tool to study composite systems~\cite{neumann27a}.
The von Neumann entropy of a random matrix, $S^{VN}$, can be formulated similarly to
the Shannon entropy as
\begin{equation}
    \centering
    S^{VN} = -tr(\rho \ln \rho),
    \label{eq:vn-entropy}
\end{equation}
where $\rho$ is a matrix with unit trace.
However, it is often more convenient to compute the entropy in terms of the eigenvalues
of $\rho$, denoted $\lambda_{i}$, as
\begin{equation}
    \centering
    S^{VN} = -\sum\limits_{i}\lambda_{i}\ln \lambda_{i},
    \label{eq:eig-entropy}
\end{equation}
where it can be seen as a proper extension to the Shannon entropy for random matrices.

In the context of covariance matrices in statistics or density matrices in quantum mechanics,
the von Neumann entropy provides a measure of correlation between states of a system~\cite{demarie18a, tsuda04a}.

In this work, the NTK matrix acts as a kernel matrix comparing the similarity of the gradients
between points in the training data-set.
The impact of these gradients on model updates are apparent when examining the work of
\citet{krippendorf22a} where the continuous time evolution of an NN was
derived as a function of the normalized eigenvalues of the NTK matrix as
\begin{equation}
    \centering
    \dot{\tilde{f}}(\mathcal{D})= {\rm diag}(\lambda_1,\ldots,\lambda_N){\cal L}'(\mathcal{D}),
    \label{eq:ntk-eig-evolution}
\end{equation}
where $\tilde{f}$ is the NN under a basis transformation.
Equation~\ref{eq:ntk-eig-evolution} frames NN model updates in such a way
that the von Neumann entropy could become a useful tool in understanding the quality of a
data-set.
Namely, a higher von Neumann entropy would suggest a more diverse update step and therefore,
perhaps a more well-trained model.
This entropy is the first of the collective variables used through this work to predict
model performance.
It should be noted that the NTK matrix does not demand unit trace, therefore, in the
entropy calculation, the eigenvalues are scaled by their sum.
Furthermore, Equation~\ref{eq:ntk-eig-evolution} highlights the role of the eigenvalue
magnitudes which will act as a scaling factor, forcing larger
update steps along specified directions.
We use this scaling as our second collective variable built from the NTK, in particular,
we use its trace,
\begin{equation}
    \centering
    Tr(\Theta) = \sum \lambda_{i} \approx \lambda_{max},
    \label{eq:trace-ntk}
\end{equation}
which turns out empirically to be well approximated by its largest eigenvalue.
We note that changes in both of these variables throughout training measures the
deviation from a constant NTK as was partially studied in \citet{krippendorf22a}.

\subsection{Random Network Distillation for Data Selection}
\label{subsec:random-network-distillation-for-data-selection}
\begin{figure}
    \centering
    \includegraphics[width=\linewidth]{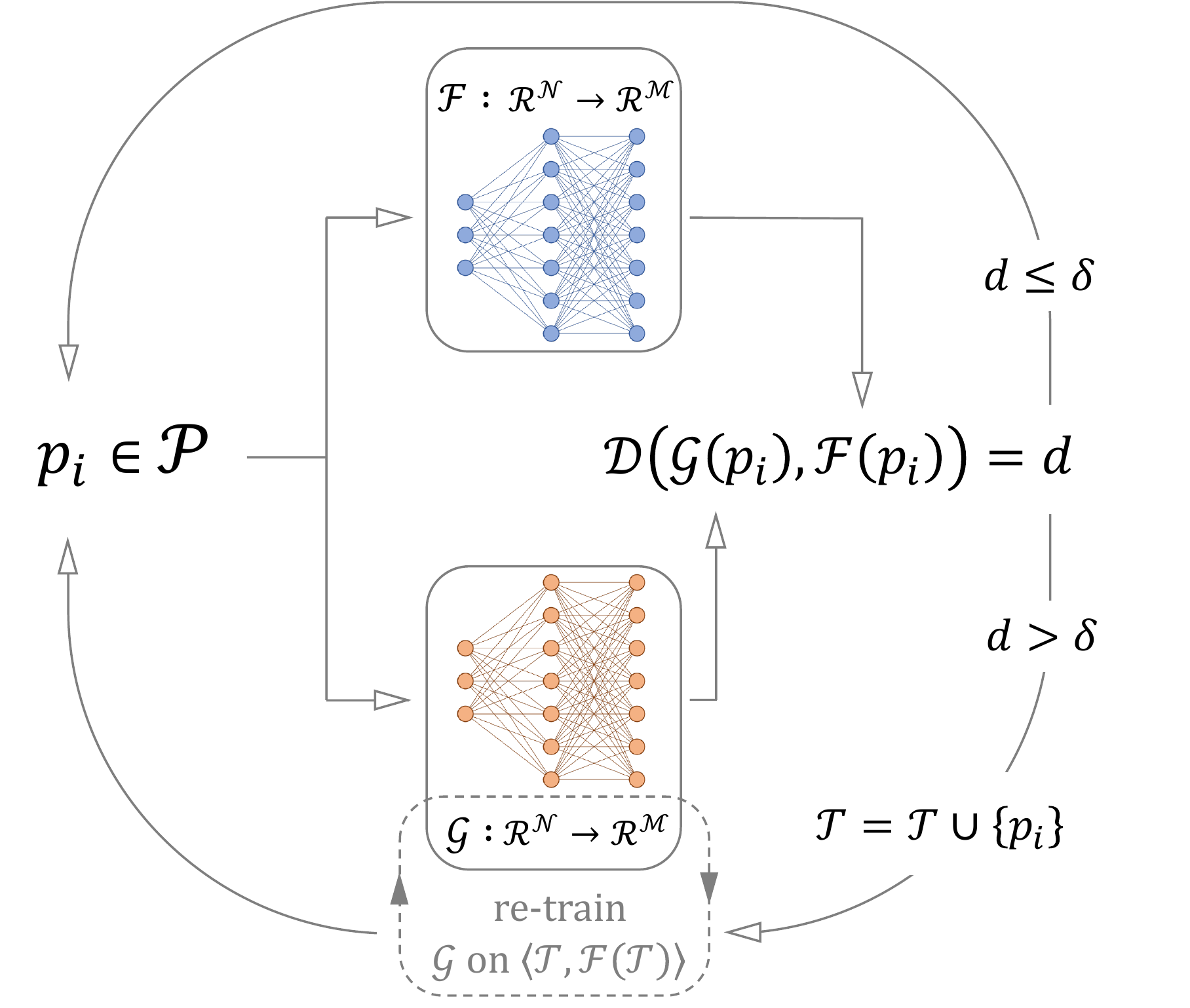}
    \caption{
    Workflow of RND.
    A data point, $p_{i}$, is passed into the target network, $\mathcal{F}$ and the predictor network $\mathcal{G}$, in order to construct the representations $\mathcal{F}(p_{i})$ and $\mathcal{G}(p_{i})$.
    A distance, $d$ is then computed using the metric $\mathcal{D}(\mathcal{F}(p_{i}), \mathcal{G}(p_{i}))$.
    If $d > \delta$, the point, $p_{i}$, will be added to the
    target set $\mathcal{T}$ and the predictor model re-trained on the full set $\mathcal{T}$.
    If the $d \leq \delta$, it is assumed that a similar point already exists in $\mathcal{T}$ and is therefore discarded.
    In our notation, $ \langle \mathcal{T}, \mathcal{F}(\mathcal{T}) \rangle $ denotes the function set with domain $\mathcal{T}$ and image $\mathcal{F}(\mathcal{T})$.
    }
    \label{fig:rnd-workflow}
\end{figure}

RND first appeared in 2018 in a paper by \citet{burda18a}
wherein the method was introduced as an approach for environment exploration in deep
reinforcement learning problems.
The concept arises from the idea that the stochastic nature of a randomly initialized
NN will act to sufficiently separate unique points from a data pool in
their high-dimensional representation space.
With this approach, it appears that an RND architecture can resolve unique
points in a sample of data such that a minimal data-set can be constructed for NN training.
The goal of this application is similar in nature to that of core-set approaches~\cite{feldman20a}
albeit using the model itself to provide information on uniqueness of training data in
an unsupervised manner.
Figure~\ref{fig:rnd-workflow} outlines graphically the process by which RND filters
points from a data pool into a target set.
The method works by randomly initializing two NNs, referred to here as the
\textit{target network} $\mathcal{F : \mathcal{R}^{N} \rightarrow \mathcal{R}^{M}}$ and the \textit{predictor network} $\mathcal{G: \mathcal{R}^{N} \rightarrow \mathcal{R}^{M}}$, which in this study are of identical architecture.
During the data selection, the target network will remain untrained while the predictor network
will be iteratively re-trained to learn the representations produced by the target network.
Theoretically, this should mean that the error between the predictor network and the
target network will provide a measure of whether a point has already been observed.
To understand this process better, we formulate it more mathematically and discuss the steps
involved individually.
Consider a set $\mathcal{P}$ consisting of points $p_{i}$ such that $i \in \mathcal{I}$
indexes $\mathcal{P}$\footnote{Indexes $\mathcal{P}$ is to say that for each
$i \in \mathcal{I}$ there exists exactly one point $p_{i} \in \mathcal{P}$.}.
Now consider a theoretical target set $\mathcal{T} \subset \mathcal{P}$ consisting of
points $t_{i}$ such that each point is maximally separated from all the others within
some tolerance $\delta$.
During each re-training run, the network $\mathcal{G}$ is trained on the elements of
$\mathcal{T}$ and target values $\mathcal{F}(t_{i} \in \mathcal{T})$.
In this way, the predictor network will effectively remember the points in $\mathcal{T}$ that it
has already seen and therefore, distinguish points from $\mathcal{P}$ that do not
resemble those already in $\mathcal{T}$.
In the case of RND for data selection, the size of $\mathcal{T}$ is set to be $S$ and points are selected for the target set in a greedy
fashion, that is, the distance between target and predictor is computed on all data points
in the point cloud and the one with the largest distance is chosen.
RND for data selection is outlined algorithmically in Algorithm~\ref{alg:rnd}.
\begin{figure}
\begin{algorithm}[H]
   \caption{Data Selection with RND}
   \label{alg:rnd}
\begin{algorithmic}
   \State {\bfseries Input:} data pool $\mathcal{P}$, target size $S$
   \While{$|\mathcal{T}| \leq S$}
   \State $D = \{d_{i} : d_{i} = \mathcal{D}(\mathcal{F}(p_{i}), {G}(p_{i}))$ $\forall$ $i \in \mathcal{P} \}$
   \State $p_{{\rm chosen}} = \{p_{i} : d_{i} \in D = \max(D)\}$
   \State $\mathcal{T} = \mathcal{T} \cup p_{{\rm chosen}}$
   \State Re-train $\mathcal{G}$ on $\langle \mathcal{T}, \mathcal{F}\left(\mathcal{T}\right) \rangle$
   \EndWhile
\end{algorithmic}
\end{algorithm}
\end{figure}
During this study, the mean square difference between representations was used as a
distance metric.

As a general note, RND is a highly involved means of data-selection and whilst the method
can be applied to the construction of data-sets consisting of hundreds of points, beyond this
will require approximation and further algorithmic improvement.
This optimization is the subject of further research and therefore, in this paper, we
construct smaller data-sets in order to better understand how they impact training.

\subsection{ZnNL}
\label{subsec:znnl}
All algorithms and workflows used in this study have been written into a Python Package called ZnNL\footnote{ZnNL can be found at https://github.com/zincware/ZnNL}.
ZnNL provides a framework for performing RND in a flexible manner on
any data as well as analyzing the selected data using the collective variables discussed
in this work.
NTK computations are handled by the neural-tangents
library~\cite{novak20a, novak22a,hron20a,sohl20a, han22a} with some additional neural
network training handled by Flax~\cite{heek20a}.
ZnNL is built on top of the Jax framework~\cite{bradbury18a} and is currently compatible with Jax-based models.

\begin{table*}[t!]
    \centering
    \caption{Table outlining the problems chosen for the experiments.
    In the case of MNIST, 10000 of the 60000 total data points were selected at random
    before the experiments took place.}
    \label{tab:data-set-table}
    \vskip 0.15in
    \begin{tabular}{@{}llllll@{}}
    \toprule
    Data-Set        & Available Data & Test Data & Problem Type   & Features & Source           \\ \midrule
    MNIST           & 10000          & 500       & Classification & 28x28x1  &~\cite{lecun98a}    \\
    Fuel Efficiency & 398            & 120       & Regression     & 8        &~\cite{quinlan93a} \\
    Gait Data       & 48             & 10        & Classification & 328      &~\cite{gumuscu19a} \\
    Concrete Data   & 103            & 10        & Regression     & 10       &~\cite{yeh07a}     \\ \bottomrule
    \end{tabular}
\end{table*}

\section{Experiments and Results}\label{sec:experiments-and-results}
In order to test the efficacy of the collective variables two experiments have been
performed.
The first investigates the correlation between the collective variables and model performance.
In the second, RND is demonstrated to outperform randomly selected data-sets before our
collective variables are used to provide an explanation for this performance.

\subsection{Investigated Data}\label{subsec:investigated-data}
To ensure a comprehensive study, several data-sets spanning  both classification and
regression ML tasks have been selected for the experiments.
To further demonstrate realistic use cases of RND as a training-set generation tool, two
of the problems have been chosen for their overall scarcity of data, making the
construction of a small, representative data-set of the utmost importance.
Table~\ref{tab:data-set-table} describes each of the chosen data-sets including
information about the ML task (classification or regression) as well as the amount
of data available and the amount used in the test sets.

\subsection{Entropy, Trace, and Model Performance}
\label{subsec:entropy-eigenvalues-and-model-performance}
In the first experiment, the correlation between our collective variables and model performance
is examined.
To do so, NNs were trained with a constant architecture but varying initialization and
training data for the MNIST and Fuel data-sets.
In addition to changing the data-set, a dense and convolutional model architecture was
tested for the MNIST classification.
Details of the experiment are summarised in Table~\ref{tab:correlation-test}.
\begin{table*}[h]
\centering
\caption{
Parameters used in the study of entropy and NTK trace with respect to model training.
Network architecture nomencalture is defined in Table~\ref{tab:config-rnd}.
ReLU activation has been used between hidden layers and an ADAM optimizer in the training.
}
\label{tab:correlation-test}
\vskip 0.15in
\begin{tabular}{@{}ccccc@{}}
\toprule
Data-set Name       & Architecture                                                                                                                       & \# Samples & Max Accuracy & Min Test Loss \\ \midrule
Fuel Dense          & $\left( \mathcal{D}^{128}, \mathcal{D}^{128}, \mathcal{D}^{1}  \right )$                                                           & 7075              & N/A              & 0.051             \\
MNIST Dense         & $\left( \mathcal{D}^{128}, \mathcal{D}^{128}, \mathcal{D}^{10}  \right )$                                                          & 5480              & 95.000           & 0.015             \\
MNIST Conv. & $\left( \mathcal{C}^{32}_{2 \times 2}, \mathcal{AP}_{2 \times 2}^{4 \times 4}, \mathcal{C}^{64}_{2 \times 2}, \mathcal{AP}_{2 \times 2}^{4 \times 4}, \mathcal{D}^{128},  \mathcal{D}^{10}  \right )$ & 3082              & 99.000           & 0.008             \\ \bottomrule
\end{tabular}
\end{table*}
In each experiment, a data-set size was randomly generated and the NN parameters randomly
initialized to sample the entropy and trace space.
The trace and entropy of the NTK were then computed at the beginning of the training process,
i.e. before the first back-propagation step.
The discussion to follow pertains to models initialized using a standard LeCun procedure~\cite{lecun12a}.
The same study has been performed for NTK initialized~\cite{novak20a} models and is presented in Appendix~\ref{subsec:entropy-trace-relations-for-ntk-parametrization}.
\begin{figure*}[h!]
    \centering
    \includegraphics[width=\linewidth]{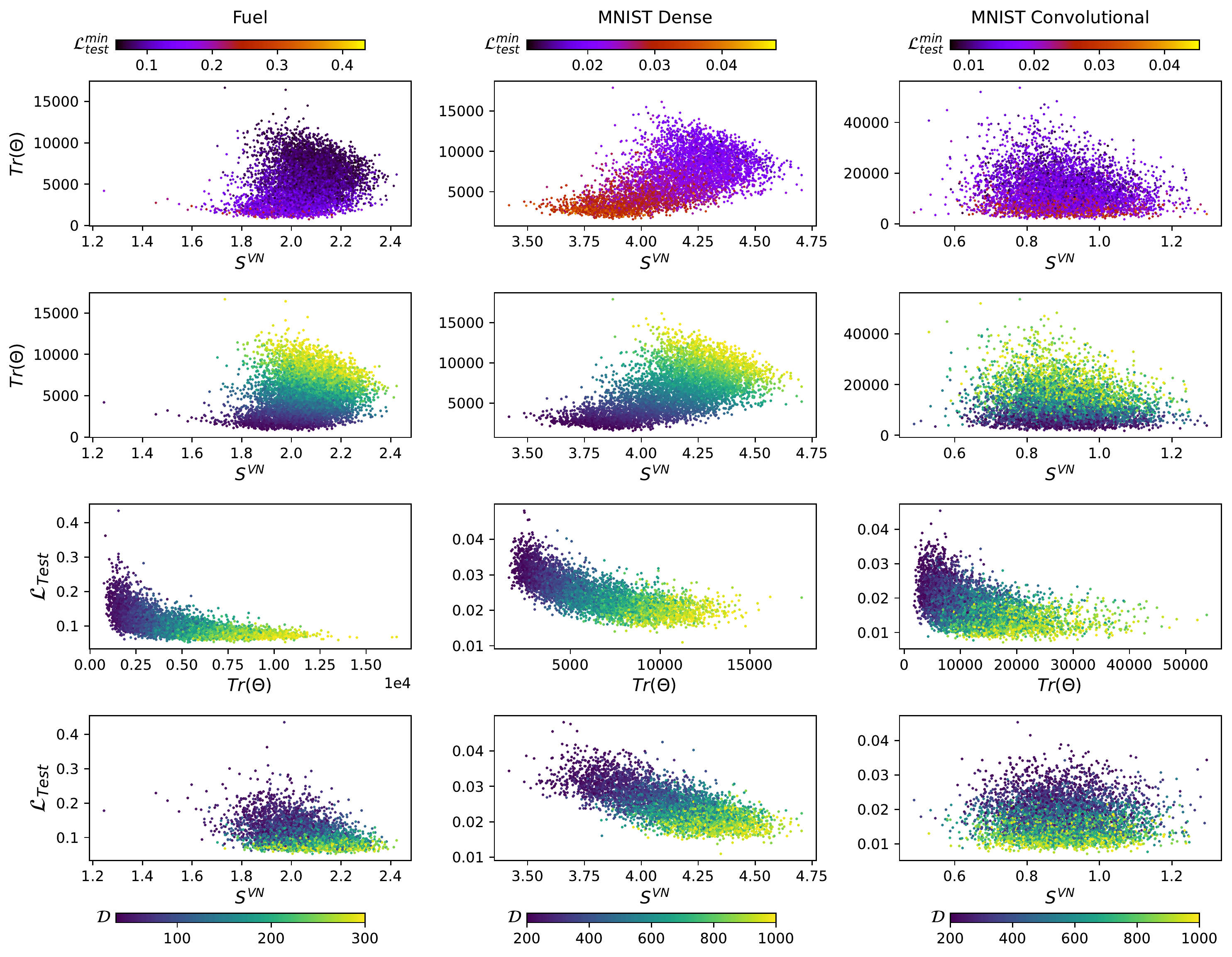}
    \caption{Figures describing the relationship between the entropy, NTK trace, and
    minimum test loss.
    Colours in the first row correspond to the minimum test loss achieved during training where a darker colour
    corresponds to a smaller loss.
    Colours in the remaining rows correspond to the size of the data-set used in the NN training with darker
    colour corresponding to smaller data-sets.}
    \label{fig:surface-data-plots}
\end{figure*}
Figure~\ref{fig:surface-data-plots} presents the outcome of this experiment with the
collective variables plotted against the minimum test loss as well as each other.
In the first row, colour corresponds to the minimum test loss achieved during training.
In the remaining rows, the colour represents the size of the data-set used in the training with darker
colours being smaller data-sets.

In the first two rows, one can see the plots of the NTK trace vs the starting entropy of the matrix.
The first of these plots is coloured by the minimum test loss achieved by the model and the second row
shows the data-set size.
What we see here for the dense models is the formation of a loss surface wherein both the
entropy and trace contribute to the performance achieved during training.
In these cases, it appears as though a combination of entropy and trace is required in
order to achieve maximal performance in model training.
In the case of the convolutional network, this trend is not as clear.
It appears that, whilst an increasing trace will aid in model performance, entropy does
not show such a clear trend.

Analysing the plots of the trace against the minimum test loss during model training,
an interesting similarity appears between the different data-sets and architectures.
Namely, the formation of a hull like shape showing decreasing test loss with increasing
starting trace.
The results suggest that a larger starting NTK trace yields models with better generalization
capacity, as demonstrated by their lower test loss.
It is notable that this trend occurs across different data-sets, architectures, and
initializations.
Secondary to the simple relationship, there also appears to be a constraint effect present.
Namely, whilst at lower trace values the models can achieve low test loss, they in general take
on a larger range of values, whereas at larger traces the spread of the values becomes slim.

Turning our attention to the entropy plots, the relationship becomes less clear.
In the dense models, a similar trend can be identified with the larger entropy data-sets
resulting in lower test loss.
However, the mechanism by which this occurs differs.
For MNIST, the larger entropy appears to fit linearly with a lower test loss, whereas in
the case of the fuel data-set, this relationship is present but slightly weaker.
What is present in both is the existence of the constrain mechanism discussed in the
trace vs entropy plots.
It appears that data-sets with larger starting entropy, no matter their size, will take on
a smaller range of minimum test losses after training.
In the case of the convolutional models, this trend is non-existent, suggesting, at least
for the tested architecture, that starting entropy is not an indicator of model performance.
It is important to note that the starting entropy and trace values will depend on the problem and
chosen architecture.
For the purpose of this study, architecture has been fixed and therefore, the effects of these
parameters is not studied and is left to future work.

In all plots there appears some degree of banding in data-set size.
Of note however is the mixing present in these bands as smaller data-sets with higher values
of the collective variables achieve test losses akin to those in the larger data-sets.
This mixing is evidence that it is the collective variables themselves and not simply
data-set size that are responsible for the results.

\begin{figure*}[h!]
    \centering
    \includegraphics[width=0.9\textwidth]{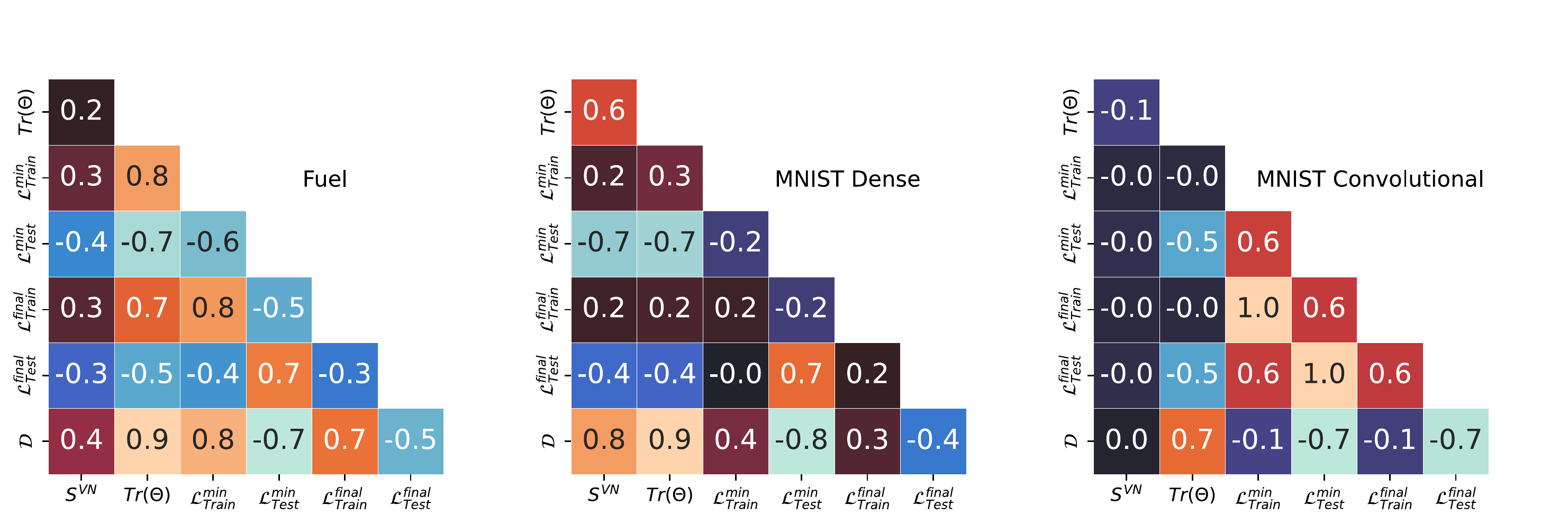}
    \caption{Correlation matrix of several variables in model training.
    Colours correspond to the numbers in the boxes.}
    \label{fig:correlation-coefficients}
\end{figure*}

Beyond the plots, the Pearson correlation coefficients between several variables have also
been computed the correlation matrices presented in Figure~\ref{fig:correlation-coefficients}.
These matrices have been constructed with additional metrics in order to present the correlation
between our collective variables and other properties of the model.
The trends discussed in the plot can be seen here numerically to correspond to our conclusions.
Relationships between the collective variables and training losses are also displayed in
these matrices.
In these cases, it appears larger values of the collective variables results in larger train
losses during training.
An explanation could be that larger entropy and trace values would correspond to fitting over
more modes in a data-set and therefore, high training losses with lower test losses.

These results highlight the correlation between the collective variables and model performance
for standard machine learning training on different data-set sizes.
In order to extend the investigation of this model, it is important to understand how
entropy changes on fixed data-set sizes can impact performance.
To this end, the efficacy of data-selection methods has been studied using these
collective variables.

\subsection{Random Network Distillation}
\label{subsec:random-network-distillation}
With the results thus-far suggesting a relationship between entropy, NTK trace,
and model performance, the remainder of the experiments will pertain to testing and interpreting
the performance of RND as a means of selecting data on which
to train.

During the RND investigations, an ensemble approach is taken in all experiments wherein the test is performed
20 times and averages of the results taken in order to construct meaningful statistics.
In this way, the stochastic initialization of the networks and the variation in data-sets
due to random selection are accounted for.
In all plots, standard error, i.e $\epsilon = \sigma / \sqrt{N}$ where $\sigma$ is the standard deviation
of the samples and $N$ is the number of samples, is shown in the error bars.

In the first part of the experiment, the efficacy of RND is assessed by
constructing data-sets of different sizes and comparing the minimum and final test losses
with data-sets constructed using random selection.
\begin{figure*}[h!]
    \centering
    \includegraphics[width=0.9\linewidth]{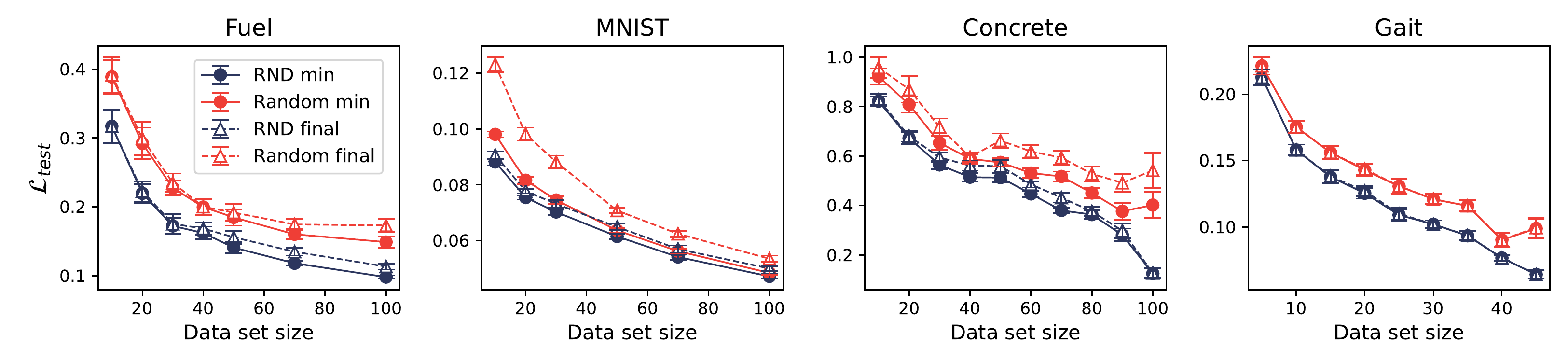}
    \caption{
    Minimum test loss and final test loss of models trained on data-sets chosen by RND and randomly
    for several data-set sizes.
    Size of the error bars corresponds to the ensemble operation over models wherein the
    experiment was performed 20 times for a single data-set size.
    }
    \label{fig:ds-size-vs-min-loss}
\end{figure*}
Figure~\ref{fig:ds-size-vs-min-loss} presents the results of this experiment.
Examining first the minimum test loss, it can be seen that data-sets generated using RND consistently outperform
those constructed using random data selection.
This is true for all data-set sizes, problems, and ML tasks, suggesting RND is a
suitable tool for optimal data-set construction.
The final test loss plot in Figure~\ref{fig:ds-size-vs-min-loss} was also compared to identify any effect on over-fitting
in the models.
The results of this comparison, show that RND selected data-sets not only provide better minimum loss
but also appear less susceptible to over-fitting.

These results suggest that RND is capable of producing a data-set that
spans the problem domain in a minimal number of points, resulting in a low minimum loss.
Furthermore, the marked reduction in over-fitting in the RND data-sets indicates that
the data used in the training covered a more diverse region of the problem space, avoiding
similar elements.

In the next part of the experiment, the starting von Neumann entropy and trace of the NTK matrix is computed
for each data-set size and a comparison between randomly selected sets and RND selected
sets is investigated.
\begin{figure*}[h!]
    \centering
    \includegraphics[width=0.9\linewidth]{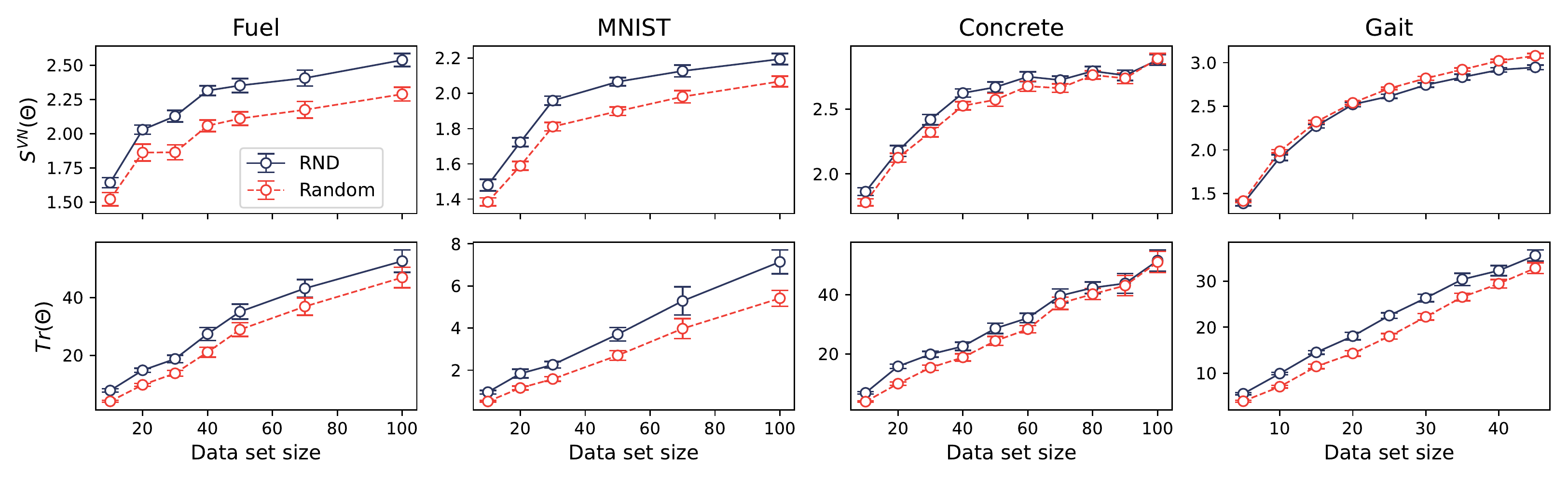}
    \caption{
    Starting von Neumann entropy and trace of the NTK matrix constructed on data-sets of different
    sizes produced with both random and RND approaches.
    Size of the error bars corresponds to the ensembling described in the text.
    }
    \label{fig:entropy-vs-ds-size}
\end{figure*}
In Figure~\ref{fig:entropy-vs-ds-size}, the results of this investigation are presented
as plots of the starting entropy and trace vs the data-set size for the different problem sets.
These plots clearly show that in each case, RND selected data-sets have a higher
starting entropy  and/or NTK trace than those selected randomly.

To understand how these variables have an impact on training, the NTK matrix must be examined more closely.
The elements of the NTK matrix describe the similarity of the gradient vectors formed by the
representation of points of a data-set in the embedding space of an NN with
respect to the current parameter state.
Consider the NTK formed by two points selected from a data-set.
If the gradient vectors computed for these points align, the inner-product will be large
suggesting that they will evolve in a similar way under a parameter update of the network.
In this case, the entropy of the NTK will be low as only one of the two points is
required to explain this evolution.
In the case where these points are almost perpendicular to one another, their inner product
will be small and their entropy high as the NTK matrix takes on the form of a kernel matrix
dominated by its on-diagonal entries.
This will mean that during a parameter update, both points will contribute in different ways to the learning.
These conclusions can be further explained with the work of \citet{krippendorf22a} wherein a model update
is written in the form of Equation~\ref{eq:ntk-eig-evolution}.
In this form, one can see that the update step along a specific eigenmode in the data will be scaled by the magnitude
of the associated eigenvalue $\lambda$.
Therefore, a larger eigenvalue will result in a larger gradient step along this mode
and ideally, better training.
Such a result recommends that the trace should be maximised in order to focus on dominant
eigenmodes and better train the model.
However, entropy maximisation would be equivalent to increasing the number of dominant
eigenmodes within the system, thereby redistributing the eigenvalues.
In this way, a balance between the number of dominant modes in the system, represented by
the entropy, and the scale factor of each mode, represented by the trace, should be achieved
for ideal model performance.

Here it has been shown that RND selected data-sets typically produce data-sets where one or both of the
trace and entropy of a data-set with respect to an NN architecture is larger than an analogous data-set chosen randomly.
Whilst it seems that there is a correlation between these variables and the model performance,
it is not clear thus-far how best to disentangle their individual roles in the model
updates and further work is needed to explore this.
Furthermore, work here has not touched upon the role of architecture in the scaling of
the collective variables.
This remains the topic of future investigations.

\section{Conclusion}\label{sec:conclusion}
This work has examined the performance of finite width NNs by studying the
spectrum and entropy of the associated NTK matrix computed on training data.
It has been shown that there exists correlation between the starting entropy and trace
of the NTK matrix and model generalisation seen after training as measured through the test loss.
These collective variables enable us to quantify the effect of different data selection methods
on test performance.
Our results support previous work performed into understanding how modes of data-sets are
learned by models, namely the relationship between larger eigenvalues and better
training.
This framework has been applied to the understanding of RND as a
data-selection method.
The efficacy of RND has been shown on several data-sets spanning regression and
classification tasks on different architectures.
In order to explain this performance, it was shown that RND selects data-sets that
have larger starting entropy and/or NTK trace than those selected randomly.
This work acts as a step towards the construction of a general, phenomenological theory of
machine learning training in terms of the collective variables of entropy and NTK trace.
Future work will revolve around further disentangling the role of entropy and trace in other aspects
of NN training including architecture and optimizer construction as well as
better understanding their evolution during training.
With the ever-growing complexity of NNs, a framework built upon physically
motivated collective variables offers a rare explainable insight into the inner-workings of
these complex models.
The work presented here is a first step in building a deeper understanding of this
framework and perhaps, will act as a platform for the construction of a comprehensive theory.

\section*{Acknowledgements}
C.H and S.T acknowledge financial support from the German Funding Agency (Deutsche Forschungsgemeinschaft DFG) under Germany’s Excellence Strategy EXC 2075-390740016, and S. T was supported by a LGF stipend of the state of Baden-W\"{u}rttemberg.
C.H, and S.T acknowledge financial support from the German Funding Agency (Deutsche Forschungsgemeinschaft DFG) under the Priority Program SPP 2363.

\bibliography{references}

\begin{thebibliography}{29}
\providecommand{\natexlab}[1]{#1}
\providecommand{\url}[1]{\texttt{#1}}
\expandafter\ifx\csname urlstyle\endcsname\relax
  \providecommand{\doi}[1]{doi: #1}\else
  \providecommand{\doi}{doi: \begingroup \urlstyle{rm}\Url}\fi

\bibitem[Bradbury et~al.(2018)Bradbury, Frostig, Hawkins, Johnson, Leary,
  Maclaurin, Necula, Paszke, Vander{P}las, Wanderman-{M}ilne, and
  Zhang]{bradbury18a}
J.~Bradbury, R.~Frostig, P.~Hawkins, M.~J. Johnson, C.~Leary, D.~Maclaurin,
  G.~Necula, A.~Paszke, J.~Vander{P}las, S.~Wanderman-{M}ilne, and Q.~Zhang.
\newblock {JAX}: composable transformations of {P}ython+{N}um{P}y programs,
  2018.

\bibitem[Burda et~al.(2018)Burda, Edwards, Storkey, and Klimov]{burda18a}
Y.~Burda, H.~Edwards, A.~Storkey, and O.~Klimov.
\newblock Exploration by random network distillation, 2018.

\bibitem[Demarie(2018)]{demarie18a}
T.~F. Demarie.
\newblock Pedagogical introduction to the entropy of entanglement for gaussian
  states.
\newblock \emph{European Journal of Physics}, 39\penalty0 (3):\penalty0 035302,
  mar 2018.
\newblock \doi{10.1088/1361-6404/aaaad0}.

\bibitem[Feldman(2020)]{feldman20a}
D.~Feldman.
\newblock Core-sets: An updated survey.
\newblock \emph{WIREs Data Mining and Knowledge Discovery}, 10\penalty0
  (1):\penalty0 e1335, 2020.
\newblock \doi{https://doi.org/10.1002/widm.1335}.

\bibitem[Gur-Ari et~al.(2018)Gur-Ari, Roberts, and Dyer]{ari18a}
G.~Gur-Ari, D.~A. Roberts, and E.~Dyer.
\newblock Gradient descent happens in a tiny subspace, 2018.

\bibitem[Gümüşçü(2019)]{gumuscu19a}
A.~Gümüşçü.
\newblock Improvement of wearable gait analysis sensor based human
  classification using feature selection algorithms.
\newblock \emph{Fırat Üniversitesi Mühendislik Bilimleri Dergisi},
  31:\penalty0 463--471, 09 2019.
\newblock \doi{10.35234/fumbd.554789}.

\bibitem[Han et~al.(2022)Han, Zandieh, Lee, Novak, Xiao, and Karbasi]{han22a}
I.~Han, A.~Zandieh, J.~Lee, R.~Novak, L.~Xiao, and A.~Karbasi.
\newblock Fast neural kernel embeddings for general activations.
\newblock In \emph{Advances in Neural Information Processing Systems}, 2022.

\bibitem[Heek et~al.(2020)Heek, Levskaya, Oliver, Ritter, Rondepierre, Steiner,
  and van {Z}ee]{heek20a}
J.~Heek, A.~Levskaya, A.~Oliver, M.~Ritter, B.~Rondepierre, A.~Steiner, and
  M.~van {Z}ee.
\newblock {F}lax: A neural network library and ecosystem for {JAX}, 2020.

\bibitem[Horn and Johnson(1990)]{horn90a}
R.~A. Horn and C.~R. Johnson.
\newblock \emph{Matrix Analysis}.
\newblock Cambridge University Press, 1990.
\newblock ISBN 0521386322.

\bibitem[Hornik et~al.(1989)Hornik, Stinchcombe, and White]{hornik89a}
K.~Hornik, M.~Stinchcombe, and H.~White.
\newblock Multilayer feedforward networks are universal approximators.
\newblock \emph{Neural Networks}, 2\penalty0 (5):\penalty0 359--366, 1989.
\newblock ISSN 0893-6080.
\newblock \doi{https://doi.org/10.1016/0893-6080(89)90020-8}.

\bibitem[Hron et~al.(2020)Hron, Bahri, Sohl-Dickstein, and Novak]{hron20a}
J.~Hron, Y.~Bahri, J.~Sohl-Dickstein, and R.~Novak.
\newblock Infinite attention: Nngp and ntk for deep attention networks.
\newblock In \emph{International Conference on Machine Learning}, 2020.

\bibitem[Jacot et~al.(2018)Jacot, Gabriel, and Hongler]{jacot18a}
A.~Jacot, F.~Gabriel, and C.~Hongler.
\newblock Neural tangent kernel: Convergence and generalization in neural
  networks, 2018.

\bibitem[Krippendorf and Spannowsky(2022)]{krippendorf22a}
S.~Krippendorf and M.~Spannowsky.
\newblock A duality connecting neural network and cosmological dynamics.
\newblock \emph{Machine Learning: Science and Technology}, 3\penalty0
  (3):\penalty0 035011, aug 2022.
\newblock \doi{10.1088/2632-2153/ac87e9}.
\newblock URL \url{https://dx.doi.org/10.1088/2632-2153/ac87e9}.

\bibitem[Lecun et~al.(1998)Lecun, Bottou, Bengio, and Haffner]{lecun98a}
Y.~Lecun, L.~Bottou, Y.~Bengio, and P.~Haffner.
\newblock Gradient-based learning applied to document recognition.
\newblock \emph{Proceedings of the IEEE}, 86\penalty0 (11):\penalty0
  2278--2324, 1998.
\newblock \doi{10.1109/5.726791}.

\bibitem[LeCun et~al.(2012)LeCun, Bottou, Orr, and M{\"u}ller]{lecun12a}
Y.~A. LeCun, L.~Bottou, G.~B. Orr, and K.-R. M{\"u}ller.
\newblock \emph{Efficient BackProp}, pages 9--48.
\newblock Springer Berlin Heidelberg, Berlin, Heidelberg, 2012.
\newblock ISBN 978-3-642-35289-8.
\newblock \doi{10.1007/978-3-642-35289-8_3}.
\newblock URL \url{https://doi.org/10.1007/978-3-642-35289-8_3}.

\bibitem[Lee et~al.(2020)Lee, Xiao, Schoenholz, Bahri, Novak, Sohl-Dickstein,
  and Pennington]{lee20a}
J.~Lee, L.~Xiao, S.~S. Schoenholz, Y.~Bahri, R.~Novak, J.~Sohl-Dickstein, and
  J.~Pennington.
\newblock Wide neural networks of any depth evolve as linear models under
  gradient descent.
\newblock \emph{Journal of Statistical Mechanics: Theory and Experiment},
  2020\penalty0 (12):\penalty0 124002, dec 2020.
\newblock \doi{10.1088/1742-5468/abc62b}.

\bibitem[Neal(1995)]{neal95a}
R.~M. Neal.
\newblock Bayesian learning for neural networks.
\newblock 1995.

\bibitem[Neumann(1927)]{neumann27a}
J.~v. Neumann.
\newblock Thermodynamik quantenmechanischer gesamtheiten.
\newblock \emph{Nachrichten von der Gesellschaft der Wissenschaften zu
  Göttingen, Mathematisch-Physikalische Klasse}, 1927:\penalty0 273--291,
  1927.
\newblock URL \url{http://eudml.org/doc/59231}.

\bibitem[Novak et~al.(2020)Novak, Xiao, Hron, Lee, Alemi, Sohl-Dickstein, and
  Schoenholz]{novak20a}
R.~Novak, L.~Xiao, J.~Hron, J.~Lee, A.~A. Alemi, J.~Sohl-Dickstein, and S.~S.
  Schoenholz.
\newblock Neural tangents: Fast and easy infinite neural networks in python.
\newblock In \emph{International Conference on Learning Representations}, 2020.

\bibitem[Novak et~al.(2022)Novak, Sohl-Dickstein, and Schoenholz]{novak22a}
R.~Novak, J.~Sohl-Dickstein, and S.~S. Schoenholz.
\newblock Fast finite width neural tangent kernel.
\newblock In \emph{International Conference on Machine Learning}, 2022.

\bibitem[Ortiz-Jimenez et~al.(2020)Ortiz-Jimenez, Modas, Moosavi-Dezfooli, and
  Frossard]{ortiz20a}
G.~Ortiz-Jimenez, A.~Modas, S.-M. Moosavi-Dezfooli, and P.~Frossard.
\newblock Neural anisotropy directions, 2020.

\bibitem[Ortiz-Jiménez et~al.(2021)Ortiz-Jiménez, Moosavi-Dezfooli, and
  Frossard]{ortiz21a}
G.~Ortiz-Jiménez, S.-M. Moosavi-Dezfooli, and P.~Frossard.
\newblock What can linearized neural networks actually say about
  generalization?, 2021.

\bibitem[Quinlan(1993)]{quinlan93a}
J.~R. Quinlan.
\newblock Combining instance-based and model-based learning.
\newblock In \emph{Proceedings of the Tenth International Conference on
  International Conference on Machine Learning}, ICML'93, page 236–243, San
  Francisco, CA, USA, 1993. Morgan Kaufmann Publishers Inc.
\newblock ISBN 1558603077.

\bibitem[Rasmussen and Williams(2005)]{rasmussen05a}
C.~E. Rasmussen and C.~K.~I. Williams.
\newblock \emph{Gaussian Processes for Machine Learning (Adaptive Computation
  and Machine Learning)}.
\newblock The MIT Press, 2005.
\newblock ISBN 026218253X.

\bibitem[Shannon(1948)]{shannon48a}
C.~E. Shannon.
\newblock A mathematical theory of communication.
\newblock \emph{The Bell System Technical Journal}, 27\penalty0 (3):\penalty0
  379--423, 1948.
\newblock \doi{10.1002/j.1538-7305.1948.tb01338.x}.

\bibitem[Sohl-Dickstein et~al.(2020)Sohl-Dickstein, Novak, Schoenholz, and
  Lee]{sohl20a}
J.~Sohl-Dickstein, R.~Novak, S.~S. Schoenholz, and J.~Lee.
\newblock On the infinite width limit of neural networks with a standard
  parameterization, 2020.

\bibitem[Tsuda and Noble(2004)]{tsuda04a}
K.~Tsuda and W.~S. Noble.
\newblock Learning kernels from biological networks by maximizing entropy.
\newblock \emph{Bioinformatics}, 20\penalty0 (suppl\_1):\penalty0 i326--i333,
  08 2004.
\newblock ISSN 1367-4803.
\newblock \doi{10.1093/bioinformatics/bth906}.

\bibitem[Yang(2019)]{yang19a}
G.~Yang.
\newblock Scaling limits of wide neural networks with weight sharing: Gaussian
  process behavior, gradient independence, and neural tangent kernel
  derivation, 2019.

\bibitem[Yeh(2007)]{yeh07a}
I.-C. Yeh.
\newblock Modeling slump flow of concrete using second-order regressions and
  artificial neural networks.
\newblock \emph{Cement and Concrete Composites}, 29\penalty0 (6):\penalty0
  474--480, 2007.
\newblock ISSN 0958-9465.
\newblock \doi{https://doi.org/10.1016/j.cemconcomp.2007.02.001}.
\newblock URL
  \url{https://www.sciencedirect.com/science/article/pii/S0958946507000261}.

\end{thebibliography}

\newpage
\appendix
\section{Appendix}\label{sec:appendix}
The appendix is split into two sections, the first discusses the parameters of the
NNs used in the RND experiments and the second shows the correlation experiments for
NTK initialized networks.

\subsection{RND}\label{subsec:rnd}
In the RND experiments, three data-sets were studied covering regression and classification
problems in order to identify whether the method could outperform the random selection of
data.
Table~\ref{tab:config-rnd} outlines the network architectures used during this study.
\begin{table*}
\centering
\caption{
Architectures used in the study of comparing data chosen by RND and random selection.
$\mathcal{D}^n$ denotes a dense layer of $n$ dimensions and $\mathcal{C}^n_{l\times k}$ a convolutional layer of $n$ output channels with a filter shape of $l \times k$. Average pooling of window shape $n \times m$ and strides $l \times k$ is denoted $\mathcal{AP}^{n \times m}_{l \times k}$.
The training of each model was performed using the ADAM optimizer and the models are initialized using the NTK initializer.
}
\label{tab:config-rnd}
\vskip 0.15in
\begin{tabular}{@{}cc@{}}
\toprule
Data-set Name       & Architecture  \\ \midrule
Fuel          & $\left( \mathcal{D}^{32}, \mathcal{D}^{32}, \mathcal{D}^{32}, \mathcal{D}^{1}  \right )$   \\

MNIST & $\left( \mathcal{C}^{32}_{3 \times 3}, \mathcal{AP}_{2 \times 2}^{2 \times 2}, \mathcal{C}^{32}_{2 \times 2}, \mathcal{AP}_{2 \times 2}^{2 \times 2}, \mathcal{D}^{128}, \mathcal{D}^{10} \right )$ \\

Gait         & $\left( \mathcal{D}^{32}, \mathcal{D}^{32}, \mathcal{D}^{16}  \right )$  \\

Concrete        & $\left( \mathcal{D}^{32}, \mathcal{D}^{32}, \mathcal{D}^{32}, \mathcal{D}^{3}  \right )$   \\
\bottomrule
\end{tabular}
\end{table*}

\subsection{Entropy, Trace Relations for NTK Parametrization}\label{subsec:entropy-trace-relations-for-ntk-parametrization}
\begin{figure*}[h!]
    \centering
    \includegraphics[width=\linewidth]{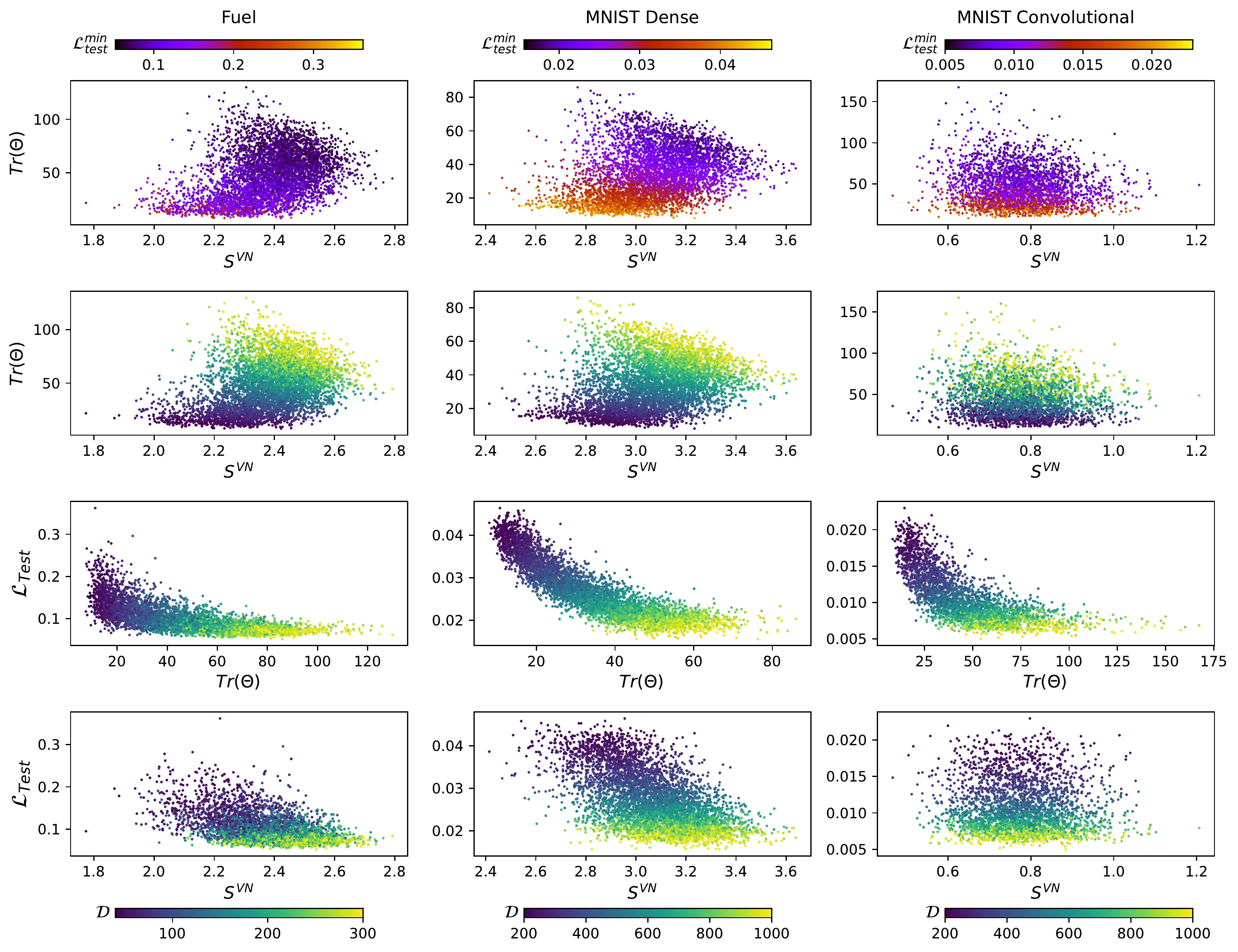}
    \caption{
    Figures describing the relationship between the entropy, NTK trace, and
    minimum test loss for NTK initialized neural networks.
    In the first row the colours correspond to the minimum test loss during training
    whereas in the remaining rows it corresponds to data-set size.
    In all cases, darker colours correspond to smaller values.
    }
    \label{fig:surface-data-plots-ntk}
\end{figure*}

\begin{figure*}[h!]
    \centering
    \includegraphics[width=\textwidth]{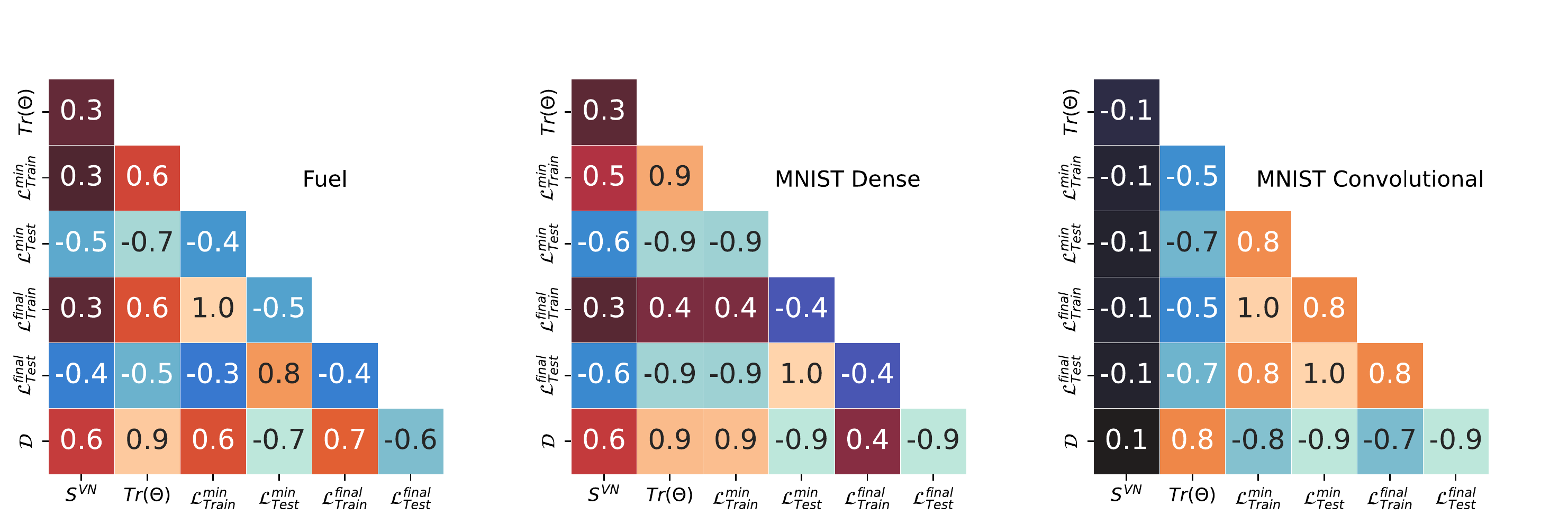}
    \caption{Correlation matrix of several variables in model training with NTK initialization.
    Colours correspond to the numbers in the boxes.}
    \label{fig:correlation-coefficients-ntk}
\end{figure*}

In order to understand the role of initialization on the introduced collective variables,
the study described in Section~\ref{subsec:entropy-eigenvalues-and-model-performance} has
also been performed using NTK initialized neural networks~\cite{novak20a}.
Figures~\ref{fig:surface-data-plots-ntk} and~\ref{fig:correlation-coefficients-ntk} detail
the results of this study.

The plot and correlation matrix share a strong similarity to those constructed under LeCun initialization, suggesting
that, at least for these two schemes, the initialization of the model did not have a large
impact on the relationships.
Interesting here is the scale of the trace, which takes on values two orders of magnitude smaller
than those in LeCun initialization.

\end{document}